\documentclass{article}

\usepackage[english]{babel}

\usepackage[letterpaper,top=2cm,bottom=2cm,left=3cm,right=3cm,marginparwidth=1.75cm]{geometry}

\usepackage{authblk}
\usepackage{amsmath}
\usepackage{graphicx}
\usepackage[colorlinks=true, allcolors=blue]{hyperref}
\usepackage{graphicx}
\usepackage{amsmath}
\usepackage{amssymb}
\usepackage{booktabs}
\usepackage{multirow}

\usepackage[capitalize]{cleveref}
\crefname{section}{Sec.}{Secs.}
\Crefname{section}{Section}{Sections}
\Crefname{table}{Table}{Tables}
\crefname{table}{Tab.}{Tabs.}

\title{Prototypical Model with Novel Information-theoretic Loss Function
for Generalized Zero Shot Learning}
\author[1]{Chunlin Ji\thanks{Corresponding author's email: chunlin.ji@gmail.com}}
\author[1]{Hanchun Shen}
\author[1]{Zhan Xiong}
\author[1]{Feng Chen}
\author[2]{Meiying Zhang}
\author[3]{Huiwen Yang}
\affil[1]{Shenzhen Origin AI Technology Co. Ltd}
\affil[2]{Department of Computer Science and Engineering, Southern University of Science and Technology}
\affil[3]{Department of Electrical Engineering and Computer Sciences, University of California, Berkeley}
\begin{document}
\maketitle

\begin{abstract}
Generalized zero shot learning (GZSL) is still a technical challenge of deep learning as it has to recognize both source and target classes without data from target classes. To preserve the semantic relation between source and target classes when only trained with data from source classes, we address the quantification of the knowledge transfer and semantic relation from an information-theoretic viewpoint. To this end, we follow the prototypical model and format the variables of concern as a probability vector. Leveraging on the proposed probability vector representation, the information measurement such as mutual information and entropy, can be effectively evaluated with simple closed forms. We discuss the choice of common embedding space and distance function when using the prototypical model. Then We propose three information-theoretic loss functions for deterministic GZSL model: a mutual information loss to bridge seen data and target classes; an uncertainty-aware entropy constraint loss to prevent overfitting when using seen data to learn the embedding of target classes; a semantic preserving cross entropy loss to preserve the semantic relation when mapping the semantic representations to the common space. Simulation shows that, as a deterministic model, our proposed method obtains state of the art results on GZSL benchmark datasets. We achieve $21\%\thicksim64\%$ improvements over the baseline model -- deep calibration network (DCN) and for the first time demonstrate a deterministic model can perform as well as generative ones. Moreover, our proposed model is compatible with generative models. Simulation studies show that by incorporating with f-CLSWGAN, we obtain comparable results compared with advanced generative models.
\end{abstract}

\section{Introduction}

Deep convolutional neural networks gain remarkable successes in object
recognition in recent years \cite{Bengio_etal_2013RL,Krizhevsky_etal_2012ICDCN,Simonyan_Zisserman_2015DCNIR}.
However, most successful deep neural networks are trained under supervised
learning frameworks, which always require a large amount of annotated
data for each class \cite{Deng2009ImageNetAL}. Inspired by human's
ability to recognize objects without having seen visual samples, recently, 
zero-shot learning (ZSL) gains a surge of interest and has been used in broad applications
\cite{Palatucci_etal_2009ZsLSO,Socher_etal_2013ZsCMT,Lampert_etal_2014AbCZsVBC,Zhang_etal_2015ZsVSSS,Xian_etal_2016LEZsC,Wu_etal_2016HOSSVU,Chao_etal_2016ZsOR,Changpinyo_etal_2016SCZs,Zhang_etal_2017LDEMZs,Zhang_etal_2018TEDEZs,Xian_etal_2018ZsCETGTBTU,Wang2019ASO}. 

ZSL offers an elegant way to extend classifiers from source categories,
of which labeled images are available during training, to target categories,
of which labeled images are not accessible.
The goal of ZSL is to recognize objects of target classes by transferring knowledge
from source classes through the relation in the semantic space, while
generalized zero-shot learning (GZSL), a more general and challenging
scenario of ZSL, tries to recognize objects from the joint set of both
source and target classes. 

Generally, methods for ZSL/GZSL can be
categorized into two majors — deterministic and generative: Deterministic methods focus on carefully designed models and semantic relation preserving the knowledge from source classes to target classes, using only the seen data from source classes; Generative methods leverage on novel generative models to transfer the knowledge of the paired relation between the semantic representation and visual feature of source classes, in order to generate the data for target classes. With these generated data, although less reliable, generative methods always obtain superior performance than deterministic methods. Broad studies show that filling the performance gap between them is a challenge. Moreover, a common problem in both methods is how to trust the less reliable data, such as using seen data of source classes to train the embedding model of target
classes or using the generated data of target classes to train the discriminative models, so that some uncertainty-aware strategies are required in such
scenarios. These two problems are the primary concerns of this study.

Two technical problems as envisioned in deterministic ZSL/GZSL \cite{Changpinyo_etal_2016SCZs,Liu_etal_2018GZsDCN}:
(i) how to bridge source classes to target classes for knowledge transfer
and (ii) how to make prediction on target classes without labeled
training data. Toward the first problem, deterministic ZSL/GZSL methods
typically embed the image features and the semantic representations
into a predefined common embedding space (with properly defined distances) using a regression model.
The choice of the embedding space and the design of regression model/neural network are essential to inherit the semantic relation meanwhile maintaining the discriminative ability. As for the second problem, we need to effectively bridge target classes to source classes by knowledge transfer such as retaining the structure of the semantic space, and prevent the overfitting in the seen data of source classes as they are blind to the semantic representations of target classes. The seminal work, deep calibration network (DCN) \cite{Liu_etal_2018GZsDCN}, introduces an entropy loss on target class which brings the semantic representations close to certain seen data of source classes. However, the entropy loss with a calibration parameter is not adequate to accurately control how much the target classes should learn from the seen data, which prevents the DCN from obtaining superior performance.

\begin{figure*}
\centering
\includegraphics[scale=0.67]{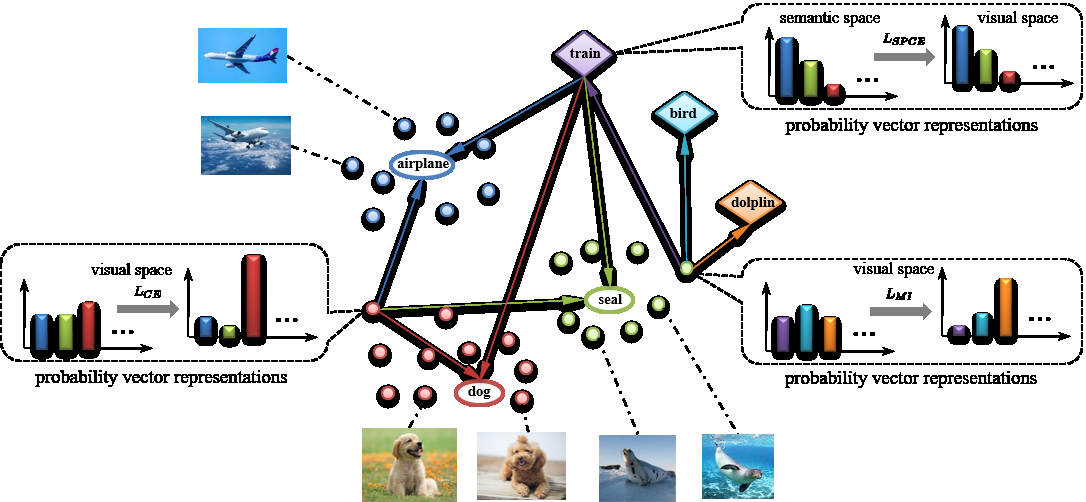}
\caption{Illustration of probability vector representation for zero-shot learning.
Circles and diamonds with text names denote the semantic representations
of all classes, small dots
denote seen data (or say visual features) of source classes, and
unseen data of target classes are unavailable. The probability vector
(PV) represents the probability that the data is assigned to different
clustering centroids/prototypes, where the assignment is shown by the arrow.
The training goal is intuitively illustrated by
the change of bars in the PV representation.}
\label{figure1}
\end{figure*}

Before the introduction of our major contributions, we address a uniform representation of the concerned variables by a soft assignment probability vector (PV), illustrated in \cref{figure1}. By regarding the semantic representation of either source
or target classes as the clustering centroids, or say some reference points in the common space, the position of visual feature in this space can be formatted into a soft assignment PV under the prototypical model \cite{Snell2017PrototypicalNF}: we characterize the position of visual feature indirectly by measuring its relation (assignment probability) to the reference points. We discuss the choice of projection model, the common space and the distance functions for the definition of the prototypical model based PV representation. Given such a uniform PV representation, we can evaluate the information measurements, such as mutual information, entropy, cross entropy with closed form expressions. 

The major contributions: 1) we propose a mutual information loss
to link the semantic representations of target classes to seen data of source classes. The mutual information consists of two parts: the conditional entropy encourages the
seen data to attach to certain prototype/centroid of the target classes, while the entropy term preserving the semantic representations collapses to trivial solutions when projecting
them to the visual space; 2) we propose an uncertainty-aware loss function which
prevents overfitting when using seen data of source classes to train the embedding model of target classes. We define a regularized entropy, which allows us to compare/control the uncertainty of the seen data belonging to source and target classes; 3) we propose
a semantic preserving loss function, which minimizes the cross entropy between the PV in original semantic space and in the visual space to preserve the semantic relation when learning the network that maps the semantic representations to the visual space.

We evaluate the performance of our proposed methods on broadly studied benchmark datasets.
Simulation shows that, as a deterministic GZSL model, our proposed
method obtains state of the art results, and outperforms significantly the recent deterministic
models on all benchmark datasets. Our proposed model is compatible with generative
models as well. We present additional loss functions to learn with generated data by considering their higher uncertainty than seen data. The experiments show that, by incorporating with generated data from f-CLSWGAN \cite{Xian_etal_2018FGNZs}, we gain obvious improvement over the vanilla f-CLSWGAN model and for the first time
demonstrate a deterministic model can perform as well as generative ones.

\section{Related works}

\textbf{Deterministic models for GZSL.} Deterministic models try to
sufficiently utilize the knowledge of the semantic embedding of both
source and target classes to conduct the inference on visual data. To this end, previous works typically embed visual samples and the semantic
embedding to a common embedding space \cite{Frome_etal_2013DEVISE,Fu_etal_2015TMvZsL,Zhang_etal_2017LDEMZs,Cacheux2019ModelingIA},
such as the visual space, the semantic embedding space or an intermediate
space between semantic and visual domains. The choice
of embedding space is critical for model performance. Previous works
\cite{Shigeto_etal_2015RSHZsL,Zhang_etal_2017LDEMZs} show that
using visual space instead of semantic space or any other intermediate
space as the common embedding space alleviates the negative effect of the hubness problem 
 \cite{Radovanovi2010HubsIS,Tomasev2014,Lazaridou2015HubnessAP}. The choice of distance function in the common embedding space also plays an importance roll. In previous studies \cite{Vinyals2016MatchingNF,Snell2017PrototypicalNF,Ravi2017OptimizationAA,Liu_etal_2018GZsDCN}, Euclidean distance, dot product similarity and cosine similarity are broadly applied.

The majority of the ZSL/GZSL methods tend to compensate the lack
of visual representation of the unseen classes with the learning of
a semantic preserving mapping. For instance, a fairly successful approach
is based on a bi-linear compatibility function that associates visual
representation and semantic features, such as ALE \cite{Akata_etal_2013LeAbC},
DEVISE \cite{Frome_etal_2013DEVISE}, SJE \cite{Akata_etal_2015EOEFGIC}
and ESZSL \cite{Paredes_Toor_2015AESATZs}. A straightforward extension
of the methods above is the exploration of a non-linear compatibility
function between visual and semantic spaces, such as a ridge regression
\cite{Shigeto_etal_2015RSHZsL}. Furthermore, in \cite{Annadani_Biswas_2018PSRZs},
they introduce explicit regularization for semantic preserving but require an extra threshold for the similarity. In another seminal work \cite{Liu_etal_2018GZsDCN}, they introduce an entropy loss to allow the embedding network of target classes trained by seen data, and a calibration parameter is required to balance the training of source classes and target classes. In our work, we introduce a series of information-theoretic loss functions which enable the use of non-linear compatibility functions. Meanwhile, these functions allow us to translate
several intuitive assumptions on the semantic relation to easy computing formulas. Moreover, we find that the conditional entropy in our mutual information loss is consistent with the entropy loss in \cite{Liu_etal_2018GZsDCN}, while the new marginal entropy term in our loss make additional effects to encourage cluster
balancing.

\textbf{Data-Generating Models for GZSL.} Generative models possess the advantage of utilizing generated image features to remove the blindness as a result of inaccessible data of target classes during training. Variational Autoencoders (VAE) \cite{Kingma_Welling_2014AEVB}
and conditional VAE \cite{Sohn2015LearningSO} based generative models
are proposed with an aim to align the visual embedding with the semantic
embedding \cite{Tsai_etal_2017LRVsE,Schonfeld_etal_2019GZsAVA,Mishra2018AGM,Keshari2020GeneralizedZL}. A VAE based algorithm can train stably, but it fails
to capture the complex distribution \cite{Bao2017CVAEGANFI}, leading
to unsatisfied results. Generative adversarial network (GAN) \cite{Goodfellow_etal_2014GAN}
has an advantage in generating more diverse data. The f-CLSWGAN \cite{Xian_etal_2018FGNZs}
is the model including a variant of an improved WGAN \cite{Arjovsky2017WassersteinG}
and a softmax classifier. f-CLSWGAN synthesizes visual features conditioned
on semantic representations, offering a shortcut directly
from a semantic descriptor to a class-conditional feature
distribution. Despite strong performance, GAN always suffers from mode collapse issues and has a unstable training phase
\cite{Arjovsky2017TowardsPM}. Fortunately, an improved deterministic
model incorporated with a generative model leads to a more advanced performance\cite{Tong2019HierarchicalDO}. In this work, we notice that the synthetic data from the generative model are generally less reliable than the seen data, so the uncertainty-aware entropy constraint loss is also applicable here. Thus, instead of constructing a complex generative model, we train the proposed model additionally with the generated data from an f-CLSWGAN, and obtain competitive results compared to recent advanced generative models.

\section{Generalized Zero-shot learning}
Following the notation in \cite{Liu_etal_2018GZsDCN}, we first present
the definition of generalized zero-shot learning as follows: suppose
we have the seen data $\mathcal{D}=\left\{ (x^{(n)},y^{(n)})\right\} _{n=1}^{N}$,
where $x^{(n)}\in\mathbb{R}^{P}$ is the feature of the $n$-th image
in the visual space $\mathbb{R}^{P}$ and $y^{(n)}\in\mathcal{S}$
is the label from the source classes $\mathcal{S}=\{1,...,S\}$. In
this study, we assume that the image feature $x$ (also named visual
embedding) has already been extracted by a pretrained deep convolutional
networks, such as ResNet \cite{He_etal_2016DRLIR}. Let $\mathcal{T}=\left\{ S+1,...,S+T\right\} $
denote the target classes, where no seen data is available in the
training phase. For each class $c\in\mathcal{S}\cup\mathcal{T}$,
let $v_{c}\in\mathbb{R}^{Q}$ denote the semantic representation in
the semantic space $\mathbb{R}^{Q}$, such as word embedding generated
by Word2Vec \cite{Mikolov_etal_2013EERVS} or visual attributes annotated
by humans to describe the visual patterns \cite{Lampert_etal_2014AbCZsVBC},
and $\mathcal{V}=\left\{ v_{c}\right\} _{c=1}^{S+T}$ denote the set
of semantic representations. In the test phase, we predict unseen
data $\mathcal{D}'=\left\{ x^{(m)}\right\} _{m=N+1}^{N+M}$ of $M$
points from either source or target classes. The task of Zero-Shot
Learning (ZSL) is that, given $\mathcal{D}$ and $\left\{ v_{c}\right\} _{c=1}^{S}$,
lean a model $\phi:x\rightarrow y$ to classify $\mathcal{D}'$ over
target classes $\mathcal{T}$. The task of Generalized Zero-Shot Learning
(GZSL) is that, given $\mathcal{D}$ and $\left\{ v_{c}\right\} _{c=1}^{S+T}$
of both source and target classes, learn a model $f:x\rightarrow y$
to classify $\mathcal{D}'$ over both source and target classes $\mathcal{S}\cup\mathcal{T}$.

\section{Proposed methods}

\subsection{Prototype model}\label{sec:protomodel}

In GSZL, to link the visual embedding in the seen data to the class
semantic representations, an intuitive way is to view the semantic representations
(or their projection in another space) as the centroids of their corresponding
classes, and learn to push the visual embedding to surround the centroid
of its belonging class. In this work, we utilize the prototypical model/networks
\cite{Snell2017PrototypicalNF}
to realize this goal. Prototypical networks learn a metric space in
which classification can be performed by computing distances between
samples and the prototype representation (or centroid) of each
class. Under the GZSL settings, we assume that the semantic representation
$v_{c}$ or its projection by a network or a liner model $\psi(v_{c})$
in a common embedding space, $\mathbb{R}^{K}$, is the prototype of
each class. For the image feature $x$, we assume a network $\phi(x)$
to transform the image feature to the same space of the prototype
$\psi(v_{c})$. Given a distance function $d:\mathbb{R}^{K}\times\mathbb{R}^{K}\rightarrow[0,+\propto)$
for measuring the distances between samples and the prototypes,
the prototype model produce a soft assignment PV, $\boldsymbol{p}=[p_{1}(y=1|x),...,p_{C}(y=C|x)]^{T}$,
over the prototypes of each classes for the data sample $x$, 
\begin{equation}\label{eq:proto}
p_{c}(y=c|x)=\frac{\exp[-d(\phi(x),\psi(v_{c}))]}{\sum_{c'}\exp[-d(\phi(x),\psi(v_{c'}))]}.
\end{equation}
Here two issues lay in the definition of the soft assignment PV $\boldsymbol{p}$,
which are the choice of predefined embedding metric space and the definition of distance function in
such space.

\subsubsection*{Choice of common embedding space}

The choice of the common embedding space is a key factor in utilizing
the prototypical model. Motivated by previous works \cite{Shigeto_etal_2015RSHZsL}\cite{Zhang_etal_2017LDEMZs},
we map the semantic representations $\mathcal{V}$ to the visual space
such that the semantic relation between the mapped semantic representations
and the visual features reflects the relation between their corresponding
classes. We propose a \emph{multilayer perceptrons (MLP)} \cite{Rumelhart_etal_1986PDP} as the compatibility function 
to map the semantic representations to the visual space $\psi:v_{c}\rightarrow z$,
where $z\in\mathbb{R}^{P}$. Therefore, the soft assignment PV $\boldsymbol{p}$
expression becomes 
\begin{equation}
p_{c}(y=c|x)=\frac{\exp[-d(x,\psi(v_{c}))]}{\sum_{c'}\exp[-d(x,\psi(v_{c'}))]}.
\end{equation}
In previous works \cite{Akata_etal_2013LeAbC,Frome_etal_2013DEVISE,Tong2019HierarchicalDO},
they use a linear mode to project the semantic representations onto
another space (visual space or common embedding space), as linear model
is easy to keep the semantic relations. However, MLP is more flexible
and can learn the nonlinear relation between the original semantic
representations and the mapping in the visual space. In
\cref{subsec:Infor_Loss}, we introduce the information-theoretic loss functions that prevail in the unreasonable nonlinear transform
for $\psi(\cdot)$. Simulation studies in \cref{subsec:Component-analysis}
verify that the choice of visual space as the common embedding space
significantly improves the performance of the proposed prototypical
model.

\subsubsection*{Choice of distance function}

The distance function $d(\cdot,\cdot)$ plays another importance role
in the prototypical model, while Euclidean distance, cosine similarity
and dot product similarity based distances have been utilized in
previous works \cite{Vinyals2016MatchingNF,Snell2017PrototypicalNF,Ravi2017OptimizationAA,Liu_etal_2018GZsDCN}. It is easy to understand that the semantic
prototypes generally contain less information than the visual embedding. So when we map the semantic representations
to the visual space, it is not appropriate to hope that the mapped
semantic representation can be well aligned to the visual feature embedding
under Euclidean distance. In comparison, cosine similarity only emphasizes
the angle between the prototypes and visual embeddings, but their
norms could be significantly different. Dot product similarity based on
distance is more flexible than cosine similarity, as it has two degrees
of freedom, such that when it is difficult to push the prototype close
to the visual embeddings of its class in the sense of maximizing the cosine
similarity, it can allow the prototype to change its norm to get larger
dot product similarity (or smaller distance). Moreover, as it will
be discussed in \cref{subsec:Infor_Loss}, we use the prototypical model to learn
the embedding of semantic representation from target classes by linking
them to seen data. In this scenario, the uncertainty of learned model
should be higher than that of learning the embedding of source classes by
seen data. To reflect this viewpoint, we propose an asymmetrical dot
produced based distance $d(x,\psi(v_{c}))=-\max\{mx\cdot\psi(v_{c}),0\}$, where $m=\mathrm{m}_{1}$ when $c\in\mathcal{S}$ and $m=\mathrm{m}_{2}$ when $c\in\mathcal{T}$. The setting of $m$ is similar to the calibration parameter $\rho$
in the DCN model \cite{Liu_etal_2018GZsDCN}, which was introduced
to balance the confidence of source classes and the uncertainty of
target classes. We observed through experiments that our model is
not sensitive to $m$, so we choose a predefined value for $m$ by
cross validation. In the simulation study (\cref{subsec:Component-analysis}),
we compared the different choice of distance functions and show the
advantage of our proposed function.

\subsubsection*{Cross entropy loss for seen data}
Given the PV expression, $\boldsymbol{p}=[p_{1}(y=1|x),...,p_{S}(y=S|x)]^{T}$
with $p_{c}(y=c|x)=\frac{\exp[-d(x,\psi(v_{c}))]}{\sum_{c'=1}^{S}\exp[-d(x,\psi(v_{c'}))]}$
for each $c\in\mathcal{S}$, and the label $y$ of the seen data from
source classes, we can define the loss function, such as cross entropy
loss, to train the prototypical model \cite{Snell2017PrototypicalNF}.
For instance, given the seen data $x^{(n)}\in\mathcal{D}$ from source
classes $\mathcal{S}$, we can learn the proposed network $\phi(\cdot)$
and $\psi(\cdot)$ by minimizing the cross entropy loss,
\begin{equation}\label{eq:ce_loss}
L_{CE}=-\frac{1}{N}\sum_{n=1}^{N}\sum_{c=1}^{S}y_{c}^{(n)}\log p_{c}(x^{(n)}).
\end{equation}
However, only the cross entropy loss is high enough to train a prototypical
model for GSZL. So, we address several novel information-theoretic loss functions to boost the performance of a deterministic prototypical model.

\subsection{Information-theoretic loss functions} \label{subsec:Infor_Loss}
In the GZSL, only the semantic embedding of target classes
is available, while the data associated are inaccessible
during training. To remove this blindness, we propose to utilize information-theoretic measurement to translate intuitive ideas of knowledge and semantic preserving into formal quantities; to bridge the source
and target classes through the seen data of source classes, we propose the
mutual information loss; to reflect the factor that the seen data should be closer to prototypes of sources classes rather than target classes, we propose an entropy constraint loss; to preserve the semantic relation of prototypes when projecting them from the original semantic space to the visual space, we propose another cross entropy loss.

\subsubsection*{Mutual information loss to link seen data and target classes}
To link the semantic embedding of target classes to visual images
of source classes, we leverage on an intuitive factor that each seen
image can be classified to the target class that is most similar
to the image's label in the source classes, rather than being classified to all target classes with equal uncertainty (or say equal assignment
probability) \cite{Liu_etal_2018GZsDCN}. Here we translate this intuitive factor into a formal information-theoretic measurement, and let the mutual information, $MI(x,c)=H(c)-H(c|x)$,
quantify the relation (or say closeness) between the seen data (visual features) and prototypes of target classes. With the prototypical model discussed in \cref{sec:protomodel}, we can obtain the probability vector that
the seed data $x$ belongs to the prototypes of target classes, $p_{c}(x)=\frac{\exp[-d(x,\psi(v_{c}))]}{\sum_{c'=S+1}^{S+T}\exp[-d(x,\psi(v_{c'}))]}$.
 To bridge the seen data and the prototypes of
target classes, we minimize the MI loss as follows,
\begin{eqnarray}\label{eq:L_MI}
L_{\mathrm{MI}}
 & = & \sum_{c}P_{c}\log P_{c}-\mathbb{E}_x\left[\sum_{c}p_{c}(x)\log p_{c}(x)\right] \nonumber \\
 & \approx &\footnotesize{\sum_{c=S+1}^{S+T}\left(\frac{1}{N}\sum_{n=1}^{N}p_{c}(x^{(n)})\right)\log\left(\frac{1}{N}\sum_{n=1}^{N}p_{c}(x^{(n)})\right)\nonumber} \\
 &  & -\frac{1}{N}\sum_{n=1}^{N}\sum_{c=S+1}^{S+T}p_{c}(x^{(n)})\log p_{c}(x^{(n)})
\end{eqnarray}
where $P_{c}=\mathbb{E}_x[p_c(x^{(n)})]\approx\frac{1}{N}\sum_{n=1}^{N}p_{c}(x^{(n)})$. $\mathbb{E}_x[\cdot]$ denotes the expectation with respective to $x$, which is always approximated by Monte Carlo approach as sample $\{x^{(n)}\}_{n=1}^N$ are available here. $P_{c}$
can be viewed as a marginal assignment probability that a sample data
belongs to the target classes. Furthermore, increasing the marginal
entropy $H(c)$ encourages cluster balancing, which is helpful for
preventing trivial solutions that map the semantic embedding of all target
classes to a prototype (or a few prototypes) in the visual
space. The second term $H(c|x)$ in \cref{eq:L_MI}, usually
named conditional entropy, measures the uncertainty that an image data
belongs to the target classes. Previous study shows that the second
term can significantly improve prediction over target classes while
having little harm on classifying seen data \cite{Liu_etal_2018GZsDCN}.
Here, we further introduce a margin for this conditional entropy, that
\begin{equation}\label{eq:conditional_entropy}\small{
   L_{\mathrm{Ent}}=\frac{1}{N}\sum_{n=1}^{N}\left[\frac{1}{\log_{2}(\textrm{\#}\mathcal{T})}\sum_{c=S+1}^{S+T}p_{c}(x^{(n)})\log p_{c}(x^{(n)})-\mathrm{margin}_1\right]^{+} }
\end{equation}
where $\textrm{\#}\mathcal{T}$ represents the number of element in
set $\mathcal{T}$, and the term $\log_{2}(\textrm{\#}\mathcal{T})$ denotes the information capacity of $\textrm{\#}\mathcal{T}$ bits. Here, we propose to regularize the entropy by dividing the information capacity term, as a consequence, the resulting \emph{regularized entropy} varies only in a small fixed-interval $(0, C_0]$, where $C_0=\log(n)/\log_2(n)$, $\forall n>1.0$ and $n\in R$. Therefore, the selection of $\mathrm{margin}_1$ becomes easy and consistent, even though the number of element in $\mathcal{T}$ varies among different datasets. We also apply the regularization for marginal entropy $H(c)$, by dividing the term $\log_{2}(\textrm{\#}\mathcal{T})$. To Finally, the improved MI loss becomes,
\begin{equation}
    \label{eq:L_MI_final}
L_{\mathrm{MI}}=\frac{1}{\log_{2}(\textrm{\#}\mathcal{T})}\sum_{c=S+1}^{S+T}P_{c}\log P_{c}-\lambda_0 L_{\mathrm{Ent}}
\end{equation}
where we introduce an additional hyperparameter $\lambda_0$ to control the contribution of conditional entropy. The value of $\lambda_0$ is selected by cross validation, which experimental is not sensitive for different datasets.

\subsubsection*{Entropy constraint loss for uncertainty-aware training}
When training the embedding network of target classes using the seen data from source classes, a notable factor is that the seen data, to a certain extension, is the out of distribution data for target classes. So, the uncertainty of classifying the seen data to target classes should be larger than that of classifying them to source classes. Here, we propose an information constraint loss to control  the entropy of seen image with respect to the
prototypes of source classes to be less than that with respect to the prototypes of target classes. We first define the entropy terms as follows,
\begin{equation*}
\mathrm{E_{u}}(x^{(n)})=\frac{1}{\log_{2}(\textrm{\#}\mathcal{T})}\sum_{c=S+1}^{S+T}p_{c}(x^{(n)})\log p_{c}(x^{(n)})
\end{equation*}
where $p_{c}(x)=\frac{\exp[-d(x,\psi(v_{c}))]}{\sum_{c'=S+1}^{S+T}\exp[-d(x,\psi(v_{c'}))]}$
is the PV that assign the seen data to each prototype of the target classes  and
\begin{equation*}
  \mathrm{E_{s}}(x^{(n)})=\frac{1}{\log_{2}(\textrm{\#}\mathcal{S})}\sum_{c=1}^{S}p_{c}(x^{(n)})\log p_{c}(x^{(n)}) 
\end{equation*}
where $p_{c}(x)=\frac{\exp[-d(x,\psi(v_{c}))]}{\sum_{c'=1}^{S}\exp[-d(x,\psi(v_{c'}))]}$
is the PV that assign the seen data to each prototype of the source classes. The entropy
constraint loss is then defined as,
\begin{equation}\label{eq:Loss_EC}
  L_{\text{EC}}=\frac{1}{N}\sum_{n=1}^{N}\left[\mathrm{E_{u}}(x^{(n)})-(\mathrm{E_{s}}(x^{(n)})+\mathrm{margin}_2)\right]^{+}  
\end{equation}
This loss reflects the expectation that the entropy $\mathrm{E_{u}}(x^{(n)})$ should be larger than the sum of $\mathrm{E_{s}}(x^{(n)})$ plus a margin $\mathrm{margin}_2$. As discussed in the last section, the regularized entropy $\mathrm{E_{u}}(x^{(n)})$ and $\mathrm{E_{s}}(x^{(n)})$ vary in the interval $(0, C_0]$, so it is not difficult to set a proper value for $\mathrm{margin}_2$.

\subsubsection*{Cross entropy loss for semantic preserving}

We map the class embeddings to the visual space such that the semantic
relation between the mapped class embeddings and the visual embedding
reflects the relation between their corresponding classes. To keep
the relation of the mapped class embedding similar to their relation
in the original semantic embedding, we introduce another novel regularization
to preserve semantic relations. We utilize the concepts of soft
assignment PV to explicitly define semantic relations between classes,
so that the objective function is specified to preserve the soft assignment
similarity between the original semantic embedding and
the mapped prototypes in visual space. By treating the source classes
embedding as the prototypes, we can assign the target class embedding.
to these prototypes, which leads to a soft assignment PV presentation. Here let  $v_i^t$ (for $i \in \left\{1,..,T \right\}$) denotes target class embedding and $v_{j}^s$ (for $j \in \left\{1,..,S \right\}$) denotes the source class embedding. The assignment
PV in the original semantic space is $p_{j}(v_{i}^{t})=\frac{\exp[-d(v_i^{t},v_{j}^{s})]}{\sum_{j'=1}^S\exp[-d(v_{i}^{t},v_{j'}^{s})]}$,
while the assignment PV after mapping to the visual space by the network
$\psi(\cdot)$ is $p_{j}^{\psi}(v_i^{t})=\frac{\exp[-d(\psi(v_i^{t}),\psi(v_{j}^{s}))]}{\sum_{j'=1}^S\exp[-d(\psi(v_i^{t}),\psi(v_{j'}^{s}))]}$.Then,
we can use cross entropy to measure the similarity between $p_{j}(v_i^{t})$
and $p_{j}^{\psi}(v_i^{t})$. The overall semantic preserving loss is
given by 
\begin{equation}
L_{\text{SPCE}}=\frac{1}{T}\sum_{i=1}^{T}\left[\frac{1}{\log_{2}(\textrm{\#}\mathcal{S})} \sum_{j=1}^{S} p_{j}(v_i^{t})\log p_{j}^{\psi}(v_i^{t})-\text{margin}_3\right]^{+}\label{eq:sp}
\end{equation}
In previous semantic preserving method \cite{Annadani_Biswas_2018PSRZs},
the difficulty is that the semantic similarity is not easy for comparison
across the original space and the mapped space. Therefore, a careful
design for the threshold on each dataset is required \cite{Annadani_Biswas_2018PSRZs}.
However, the value of regularized entropy in $L_{\text{SPCE}}$ varies only in the interval $(0, C_0]$, so it becomes easy to set a proper margin for semantic preserving.

\subsection{Learning and inference}
We combine all the four
loss functions with different weights $\lambda_{1:3}$. Therefore, we optimize
the parameters of our model by jointly learning the following loss
functions, 
\begin{equation} 
L_{D}=L_{\mathrm{CE}}+\lambda_{1}L_{\mathrm{MI}}+\lambda_{2}L_{\mathrm{EC}}+\lambda_{3}L_{\mathrm{SPCE}}\label{eq:Loss}
\end{equation}
We observed through experiments that our model is sensitive to $\lambda_{1}$
but less sensitive to $\lambda_{2}$ and $\lambda_{3}$. So we use cross validation to
set $\lambda_{1}$ for each dataset and set consistent
values of $\lambda_{2}$ and $\lambda_{3}$ for all simulation experiments except few exceptions. The network parameters
in $\psi(\cdot)$ can be efficiently optimized by SGD or Adam algorithm
with auto-differentiation technique supported in PyTorch \cite{Paszke2017AutomaticDI}.

In the test stage, the predicted class $y(x^{(n)})$ of image feature
$x^{(n)}$ is given by $y(x^{(n)})=\textrm{argmax}p_{c}(x^{(n)})$,
where $p_{c}(x^{(n)})=\frac{\exp[-d(x,\psi(v_{c}))]}{\sum_{c'}\exp[-d(x,\psi(v_{c'}))]}$
and $\psi(\cdot)$ is the trained network that maps semantic embedding
to the visual feature space. So, the prediction is made over both source
and target classes, as $c\in\mathcal{S\cup T}$ in generalized zero-shot
learning. In the conventional zero-shot learning, we only need the prediction over the target classes $c\in\mathcal{T}$.

\subsection{Cooperate with generative model}\label{sec:generative}
Most works of generative methods emphasize on the development of a sophisticated model to generate more `realistic' data for target classes. However, effective utilization of generated data is still largely ignored. We noticed that the synthetic data from the generative model are generally less reliable than the seen data, so we propose an uncertainty-aware entropy constraint to select the generated data when they are applied in training the discriminative model. Specially, we put the generated data $\{\widetilde{x}^{(m)}\}_{m=1}^{M}$ into the prototypical model where the prototypes are from target classes, and obtain the PV,  $p_{c}(\widetilde{x}^{(m)})=\frac{\exp[-d(\widetilde{x}^{(m)},\psi(v_{c}))]}{\sum_{c'=S+1}^{S+T}\exp[-d(\widetilde{x}^{(m)},\psi(v_{c'}))]}$. Then, we define the uncertainty of the generated data by the regularized entropy, $\mathrm{\widetilde{E}_{u}}(\widetilde{x}^{(m)})=\frac{1}{\log_{2}(\textrm{\#}\mathcal{T})}\sum_{c=S+1}^{S+T}p_{c}(\widetilde{x}^{(m)})\log p_{c}(\widetilde{x}^{(m)})$.
After that, we select the generated data by the criterion that  $\mathrm{\widetilde{E}_{u}}(\widetilde{x}^{(m)})<\mathrm{margin}_4$ with a predefined threshold $\mathrm{margin}_4$. This uncertainty based selection can prevent improper generated data from making a negative effect on the prediction of target classes. Let $\widetilde{x}_{sel}=\{\widetilde{x}^{(m)}_{sel}\}_{1}^{M_s}$ denote the selected generated data, which we use to train the embedding network of target classes by 
\begin{equation}
\widetilde{L}_{\mathrm{CE}}(\widetilde{x})=-\frac{1}{M_s}\sum_{m=1}^{M_s}\sum_{c=S+1}^{S+T}\widetilde{y}_{c}^{(m)}\log p_{c}(\widetilde{x}_{sel}^{(m)}),\label{eq:G_ce_loss}
\end{equation}
where the label $\widetilde{y}_{c}$ is known in the generation of the data. Let $\gamma_1$ denote the weight for this entropy loss. 

Moreover, here, we also allow the generated data to train the embedding network of source classes. To this end, we define another mutual information loss as \cref{eq:L_MI}, the difference is that the prototypes change from target classes to source classes, $p_{c}(\widetilde{x})=\frac{\exp[-d(\widetilde{x},\psi(v_{c}))]}{\sum_{c'=1}^{S}\exp[-d(\widetilde{x},\psi(v_{c'}))]}$. Let $\widetilde{L}_{\mathrm{MI}}(\widetilde{x})$ denote this mutual information loss for generated data, and $\gamma_2$ denote its weight.

Finally, putting all the loss functions together, we obtain the loss function to train the proposed model with both seen data $x$ and generated data $\widetilde{x}$, 
\begin{equation}
L_{G}=L_D(x)+\gamma_1 \widetilde{L}_{\mathrm{CE}}(\widetilde{x})+\gamma_{2}\widetilde{L}_{\mathrm{MI}}(\widetilde{x}).
\end{equation}

\section{Experiments }
We perform extensive evaluation for both conventional ZSL and
GZSL on standard benchmark datasets: AwA1, AwA2, CUB, SUN, and aPy.

\subsection{Experimental settings}
\textbf{Datasets}. The benchmark datasets are briefly described as
follow: Animals with Attributes (AwA1) \cite{Lampert_etal_2014AbCZsVBC}
is a widely-used dataset for coarse-grained zero-shot learning, which
contains 30,475 images from 50 different animal classes. A standard
split into 40 source classes and 10 target classes is provided in
\cite{Lampert_etal_2014AbCZsVBC}. A variant of this dataset is Animal
with Attributes2 (AwA2) \cite{Xian_etal_2017ZsTGTBTU} which has the
same 50 classes as AwA1, but AwA2 has 37,322 images in all which don't
overlap with images in AwA1. Caltech-UCSD-Birds-200-2011 (CUB) \cite{Wah_2011_TCuBD}
is a fine-grained dataset with large number of classes and attributes,
containing 11,788 images from 200 different types of birds annotated
with 312 attributes. The split of CUB with 150 source classes and
50 target classes is provided in \cite{Akata_etal_2016LeIC}. SUN
Attribute (SUN) \cite{Patterson_Hays_2012SAD} is another fine-grained
dataset, containing 14,340 images from 717 types of scenes annotated
with 102 attributes. The split of SUN with 645 source classes and
72 target classes is provided in \cite{Lampert_etal_2014AbCZsVBC}.
Attribute Pascal and Yahoo (aPY) \cite{Farhadi_etal_2009DOTA} is
a small-scale dataset with 64 attributes and 32 classes(20 Pascal
classes as source classes and 12 Yahoo classes as target classes).

\textbf{Image features.} Due to variations in image features used
by different zero-shot learning methods, for a fair comparison, we use
the widely-used features: 2048-dimensional ResNet-101 features provided
by \cite{Xian_etal_2018ZsCETGTBTU}. Classification accuracies of
existing methods are directly reported from their papers.

\textbf{Semantic representations.} We use the per-class continuous attributes provided with
the datasets of aPY, AwA, CUB and SUN. Note that we can also use the
Word2Vec representations as class embeddings \cite{Mikolov_etal_2013EERVS}.

\subsection{Implementation Details}
The compatibility function in the prototypical model is implemented as MLP. The input dimension
of attribute embedding is dependent on the problem. The MLP has 2
fully connected layers with 2048 hidden units. We use LeakyReLU as
the nonlinear activation function, Dropout function, for the first
layer, and Tanh for the output layer to squash the predicted values
within $[-1,1]$. The setting of the hyperparameters is given
as follows: by cross validation, we set $m_{1}=0.5$ and $m_{2}=1.0$
for the asymmetric dot product distance.
In the overall loss function \cref{eq:Loss}, we set the value
of $\lambda_{2}$ as 0.5 for all dataset, we
set the value of $\lambda_{3}$ as 0.05 for all datasets; while the value of $\lambda_{1}$ is depended on the
data sets, we choose a value between $0.025\sim1$ by cross validation,
but by observation, for the value of $\lambda_{1}$ is also relative
robust 0.025 and 0.05 is good enough for dataset AwA1/2, aPY, CUB,
and only SUN require a larger value 0.5. The margin value is chose
as follows: $\text{margin}_{1}=0.15$, $\text{margin}_{2}=0.05$,
$\text{margin}_{3}=0.3$ for dataset aPY, CUB and SUN, while $\text{margin}_{2}=0.0$
for AwA1 and AwA2. The batch size of visual feature data is set to
$512$. For the optimization, we use Adam optimizer \cite{Kingma2015AdamAM}
with constant learning rate $0.001$ and early stopping on the validation
set.

Following the Proposed Split in the Rigorous Protocol \cite{Xian_etal_2017ZsTGTBTU},
we compare three accuracies: $ACC_{ts}$, accuracy of all unseen images
in target classes; $ACC_{tr}$, accuracy of some seen images from
source classes which are not used for training. Then we compute the
harmonic mean of the two accuracies as $ACC_{H}=2(ACC_{ts}*ACC_{tr})/(ACC_{ts}+ACC_{tr})$,
which is used as the final criterion to favor high accuracies on both
source and target classes.

\subsection{Component analysis}
\label{subsec:Component-analysis}
\subsubsection*{Effectiveness of the proposed loss functions}
We illustrate how the proposed losses affect the model by three straightforward experiments. We perform the experiments on AwA1 dataset.

\begin{figure}
\centering
\includegraphics[scale=0.475]{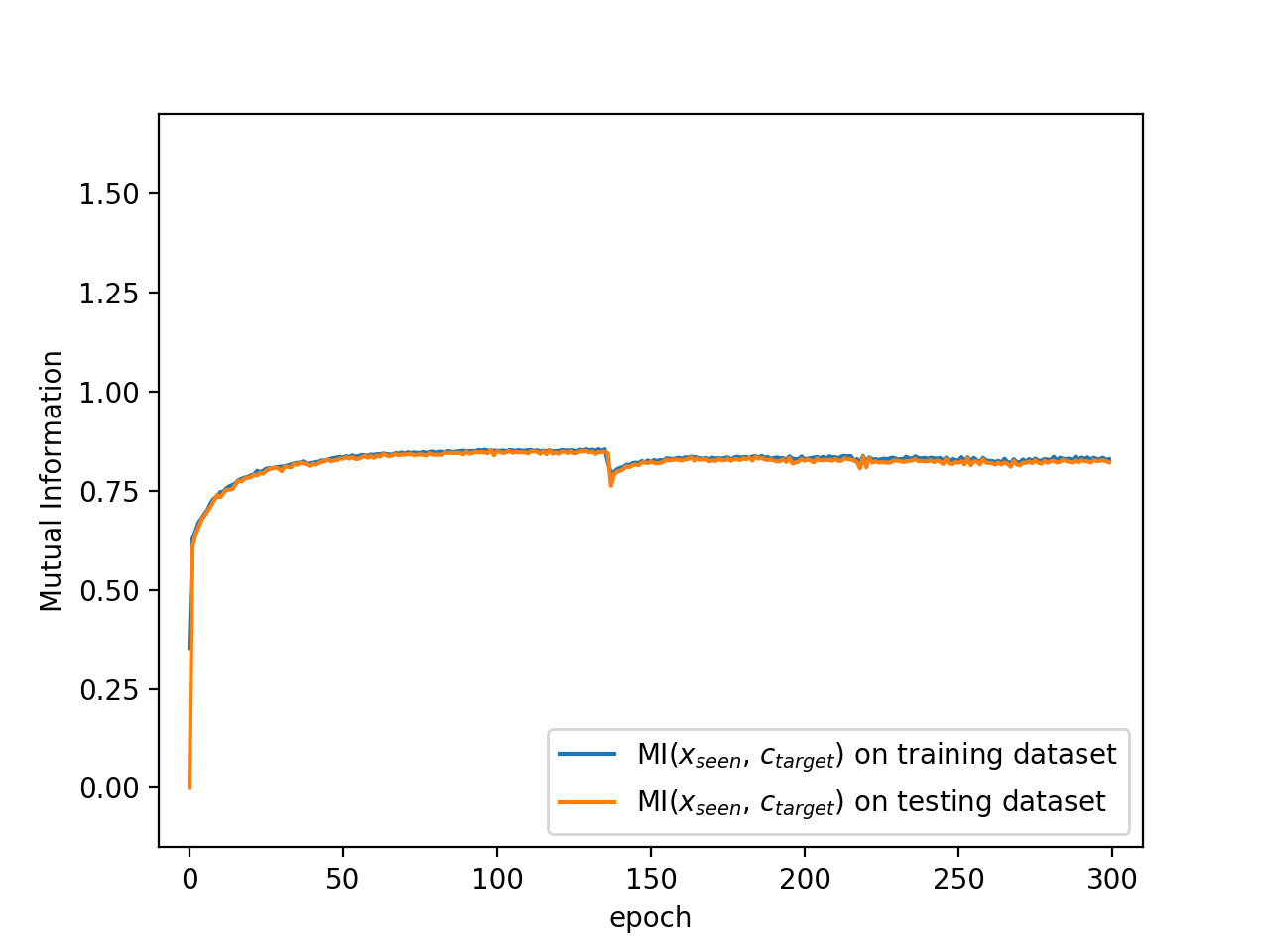}
\includegraphics[scale=0.475]{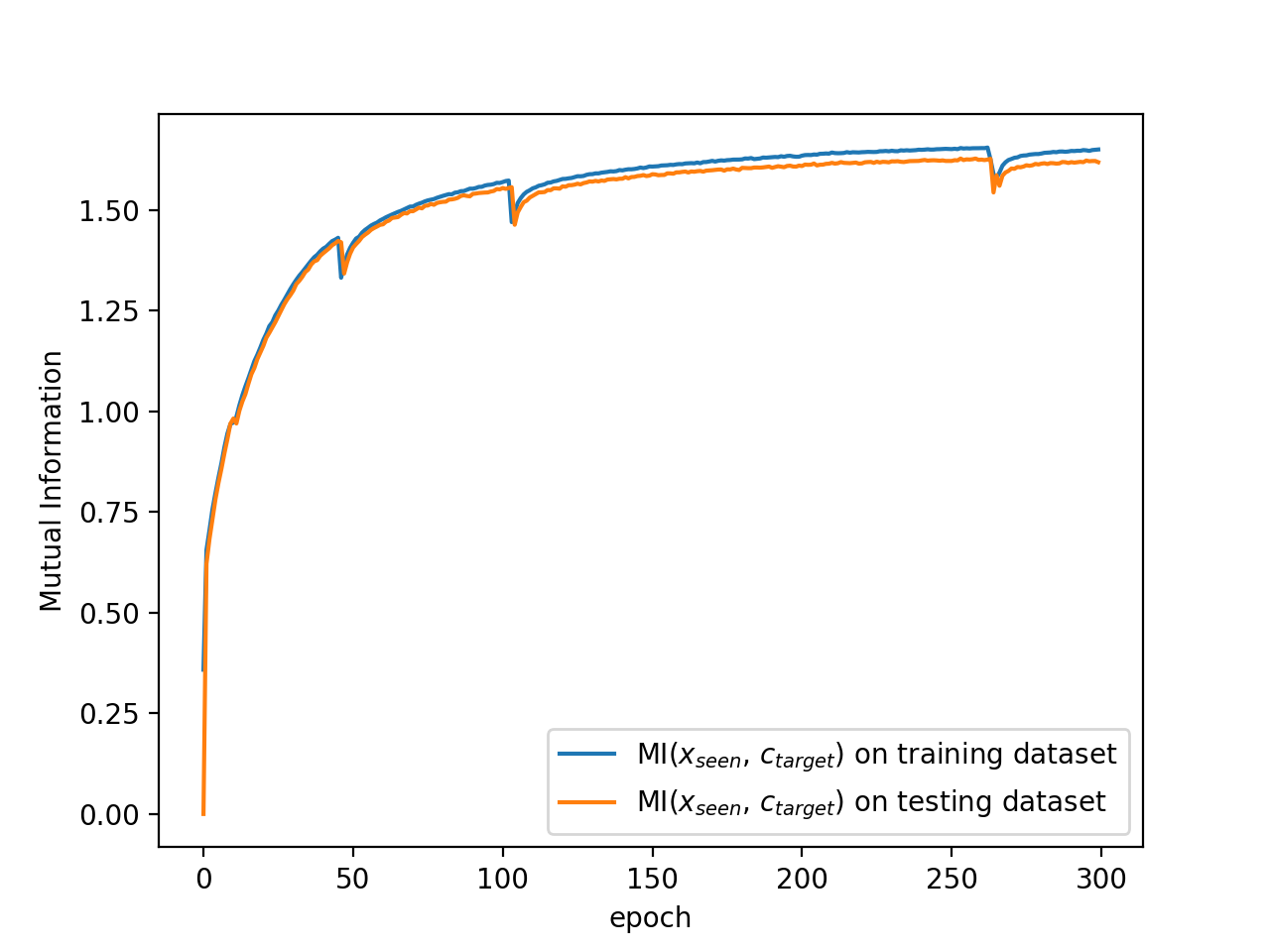}
\caption{Trajectory of mutual information $\mathrm{MI}(x_{seen},c_{target})$: Left, the model is trained without the mutual information loss $L_{\mathrm{MI}}$; Right, the model is trained with the mutual information loss $L_{\mathrm{MI}}$.}\label{figure1_MI}
\end{figure}

\textbf{Mutual information loss}: \cref{figure1_MI} shows the effectiveness of the mutual information loss function by comparing two cases: case 1 (as shown in left part of \cref{figure1_MI}), the deterministic model is trained without the mutual information loss $L_{\mathrm{MI}}$, obtaining the trajectory of $\mathrm{MI}(x_{seen},c_{target})$, the mutual information between the seen data and the labels of target classes, on both the AwA1 training and testing datasets; case 2 (as shown in right part of \cref{figure1_MI}), the deterministic model is trained with the mutual information loss $L_{\mathrm{MI}}$, getting the trajectory of $\mathrm{MI}(x_{seen},c_{target})$. From the comparison, it is noted that the $L_{\mathrm{MI}}$ loss significantly improves the mutual information between the seen data and the labels of target classes, which also means an effective knowledge/information transfer from the seen data of source classes to the classification of target classes.

\begin{figure}
\centering
\includegraphics[scale=0.475]{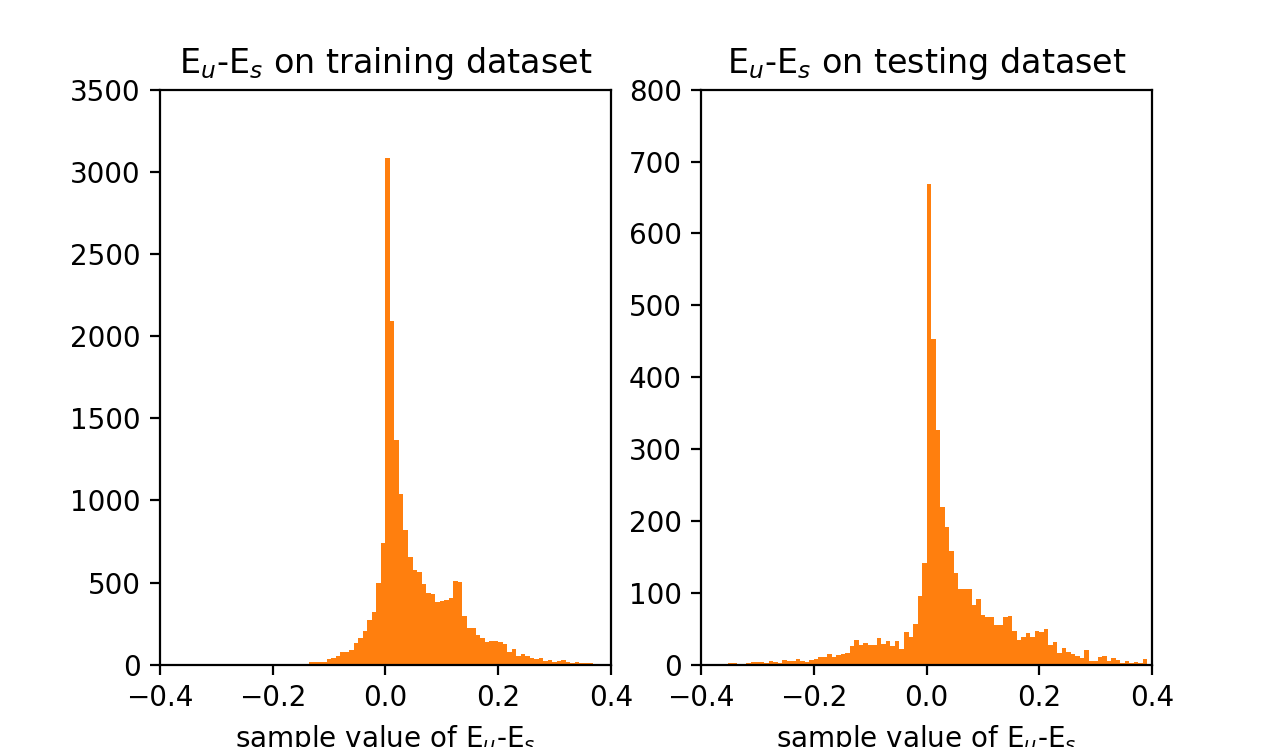}
\includegraphics[scale=0.475]{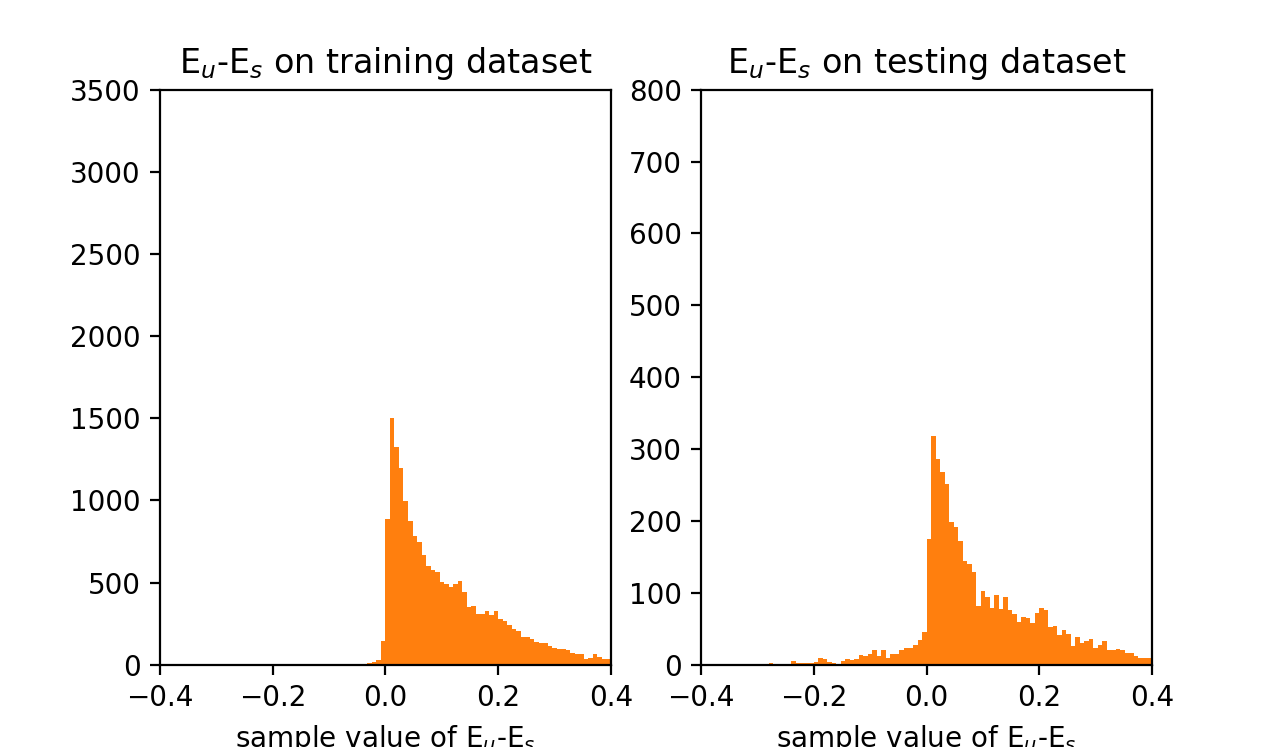}
\caption{Sample histogram of the term $E_u-E_s$: Left, the model is trained without the entropy constraint loss $L_{\mathrm{EC}}$; Right, the model is trained with entropy constraint loss $L_{\mathrm{EC}}$.}\label{figure2_EC}
\end{figure}

\textbf{Entropy constraint loss}: \cref{figure2_EC} illustrates the effectiveness of the entropy constraint loss $L_{\mathrm{EC}}$, by comparing two cases: case 1, the deterministic model is trained without this loss, showing the sample histogram of the term $E_u(x_{seen},c_{target})-E_s(x_{seen},c_{source})$ (also denoted as $E_u-E_s$), that is $\frac{1}{\log_{2}(\textrm{\#}\mathcal{T})}\sum_{c\in \mathcal{T}} p_c(y=c|x^{(n)})\log (p_c(y=c|x^{(n)}))-\frac{1}{\log_{2}(\textrm{\#}\mathcal{S})}\sum_{c \in \mathcal{S}} p_c(y^{(n)}=c|x^{(n)})\log (p_c(y^{(n)}=c|x^{(n)}))$ with $(x^{(n)}, y^{(n)})$ from the AwA1 training or testing dataset in left part of \cref{figure2_EC}; case 2, the deterministic model is trained with the entropy constraint loss $L_{\mathrm{EC}}$, the histogram as shown in right part of \cref{figure2_EC}.  In case 1, a certain percentage of the samples of  $E_u-E_s$ is negative or close to zero. However in case 2, the $L_{EC}$ loss enforces the samples of $E_u-E_s$ being larger than zero, which means the uncertainty that the seen data from source classes are classified to target classes should be larger than being classified to source classes. This comparison also shows that the entropy constraint loss mitigates the overfitting ( the seen data samples from source classes with negative $E_u-E_s$ tend to be incorrectly classified to target classes), when using seen data from source classes to train the embedding of semantic representations of target classes.

\begin{figure}
\centering
\includegraphics[scale=0.475]{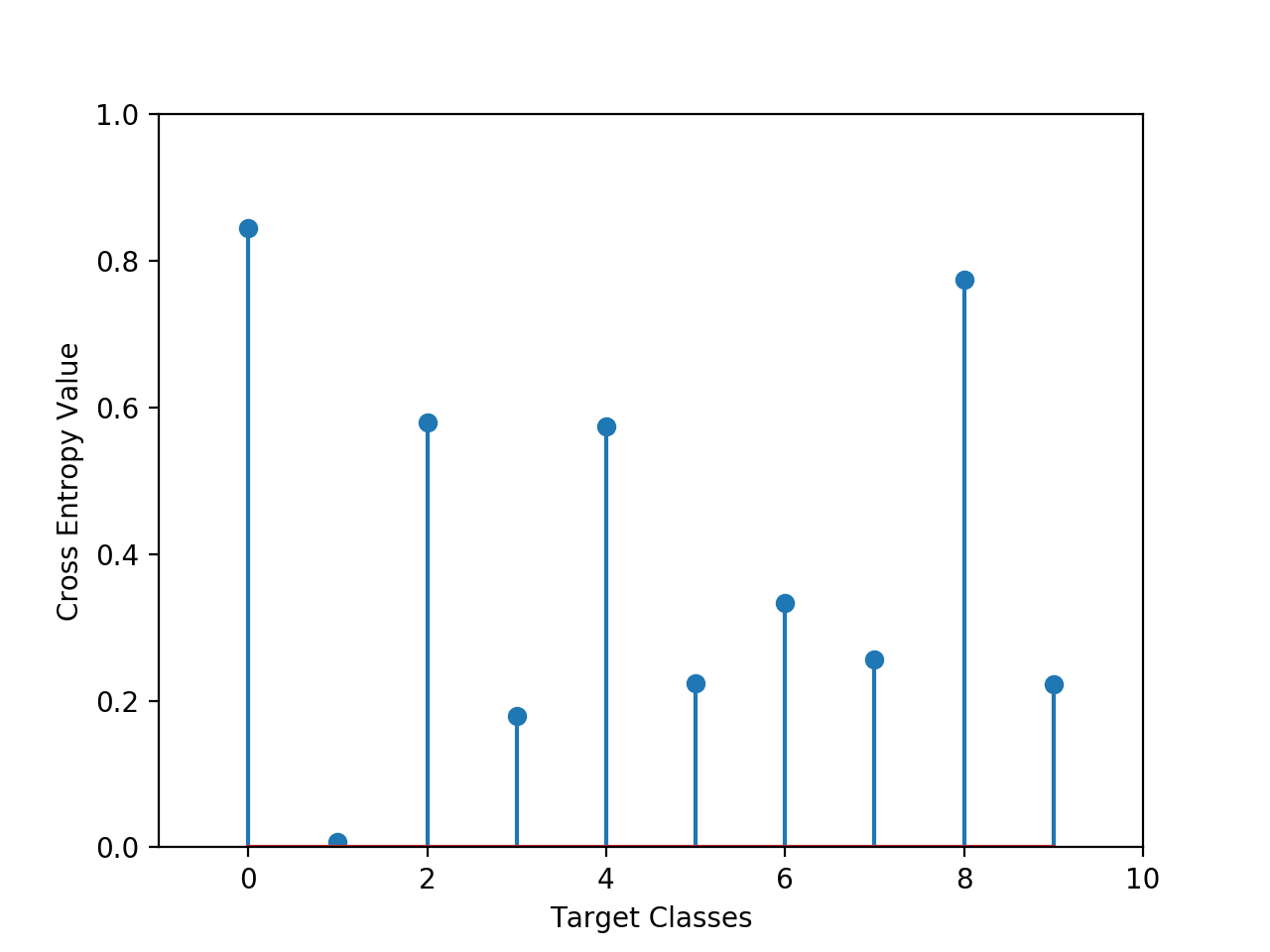}
\includegraphics[scale=0.475]{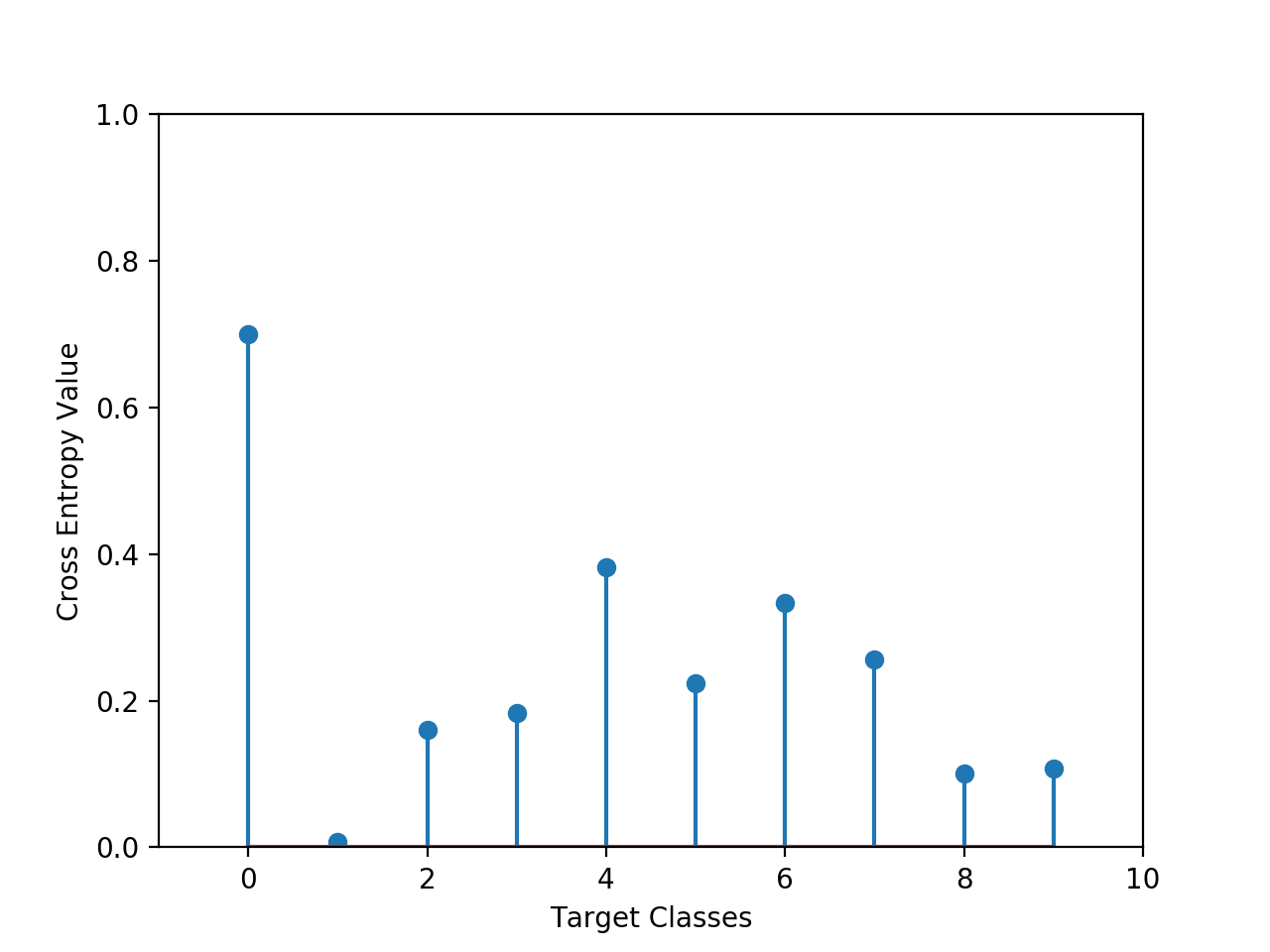}
\caption{Cross entropy between PV representations of ten target classes representations $v_{i}^t$ (for $i \in \{1,...,10\}$) in the AwA1 dataset with respect to its forty source classes representations $v_{1:40}^s$, $CE(v_i^t, v^s_{1:40},\psi)$: Left, the model is trained without the semantic preserving cross entropy loss $L_{\mathrm{SPCE}}$; Right, the model is trained with loss $L_{\mathrm{SPCE}}$.}\label{figure3_SPCE}
\end{figure}

\textbf{Semantic preserving loss}: \cref{figure3_SPCE} illustrates the effectiveness of the semantic preserving cross entropy loss $L_{\mathrm{SPCE}}$, also by comparing two cases: case 1, the deterministic model is trained without $L_{\mathrm{SPCE}}$ on the AwA1 dataset. We evaluate two PV representations that the PV for target class representation $v_i^{t}$ in the original semantic space with respect to prototypes of source classes representation $v_{j}^{s}$, $p_{j}(v_i^{t})$, and the PV after mapping them to the visual space by the network $\psi(\cdot)$, $p_{j}^{\psi}(v_i^{t})$. Then,
we utilize the cross entropy to measure the similarity between $p_{j}(v_i^{t})$
and $p_{j}^{\psi}(v_i^{t})$ for each $v_i^{t}$, that  $CE(v_i^t, v^s_{1:S},\psi)=\frac{1}{\log_{2}(S)}\sum_{j=1}^S p_{j}(v_i^{t})\log p_{j}^{\psi}(v_i^{t})$. In the left part of \cref{figure3_SPCE}, we evaluate and plot $CE(v_i^t, v^s_{1:S},\psi)$ for each $v_i^t$, where $\psi$ is the trained projection network; case 2, the deterministic model is trained with the semantic preserving cross entropy $L_{\mathrm{SPCE}}$, where we also evaluate and show the $CE(v_i^t, v^s_{1:S},\psi)$ in the right part of \cref{figure3_SPCE}.  From the comparison, it is obvious that with the semantic preserving loss $L_{\mathrm{SPCE}}$, the term $CE(v_i^t, v^s_{1:S},\psi)$ become smaller, which means that the semantic relation between the target class representations to the source class representation is more similar (or say  more inherited) after mapping these representations from the original semantic space to the visual space. Therefore, the loss $L_{\mathrm{SPCE}}$ helps to preserve the semantic relation.

\subsubsection*{Numerical evaluation on performance}
We investigate contribution of our proposed approach
on the model performance for GZSL. Here we use both the AwA1 and AwA2 datasets. We include
the result of DCN model \cite{Liu_etal_2018GZsDCN} on AwA2 and the prototypical
model trained with only the cross entropy loss as baseline methods. We
compare the choice of common embedding spaces, attribute space and feature
space, and the choice of distances, cosine and dot product. We represent the combinations as follows: space A uses the attribute space as the embedding space and uses dot product similarity based distance; space B uses the visual space as the embedding space and uses cosine similarity based distance; space C uses the visual space as the embedding space and uses dot product similarity based distance. Notice that the last two rows in \cref{all-tabel1} also choose space C. Furthermore,
we show the contribution of the proposed information-theoretic losses:
$L_{\text{Ent}}$, $L_{\text{MI}}$, $L_{\text{EC}}$ and $L_{\text{SPCE}}$. 

All the simulation results
are shown in \cref{all-tabel1}, where we cite the result of DCN directly from
\cite{Liu_etal_2018GZsDCN}. The DCN model introduces an entropy regularization for bridging seen data
and target classes, which is similar with the $L_{\text{Ent}}$ (notice that we have an extra term $\mathrm{margin}_1$ in the \cref{eq:conditional_entropy}). The DCN uses the dot product distance, and project the feature and attribute onto a common space. The third to the fifth rows show that the entropy loss $L_{\text{Ent}}$ significantly improves the GZSL performance, comparing to the cross entropy loss $L_{\text{CE}}$. And the results of the third to the fifth rows outperform the DCN significantly, which might be due to the factors that it is easier to train the model using the original attribute/feature space rather than looking for a common space, and the proposed entropy loss seems more effective than the entropy regularization in DCN. Furthermore, the third to the fifth rows demonstrate the importance of choosing the common embedding space and distance function: using the visual feature space rather than the semantic space as the embedding space gains strong improvement; using dot product similarity based distance yields better performances than using cosine similarity based distance. Comparing the sixth row with the fifth row, we show that the proposed marginal entropy $H(c)$ in $L_{\text{MI}}$ brings additional improvements. Comparing the results in the seventh row and sixth row, we show that the entropy constraint $L_{\text{EC}}$ can significantly boost the model performance. In the last row, we show that the semantic preserving loss $L_{\text{SPCE}}$ achieves a favorable improvement on AwA1 dataset but brings a negligible negative effect on AwA2 dataset. Experiments show that, for other datasets, the semantic
preserving loss also make positive contributions, so we retain
this loss in following simulation studies.

\begin{table}
\centering
\caption{Comparison of the contribution of different improvement approaches.}
\centering%
\begin{tabular}{c|ccc|ccc}
\hline 
\multirow{2}{1cm}{\centering Methods} & \multicolumn{3}{c|}{AwA1} & \multicolumn{3}{c}{AwA2}\tabularnewline
\cline{2-7}
~  & ts  & tr  & H  & ts  & tr  & H\tabularnewline
\hline
\small{DCN\cite{Liu_etal_2018GZsDCN}} & -  & -  & -  & 25.5  & 84.2  & 39.1\tabularnewline
\small{$L_{\text{CE}}$ (space C)} & 11.4  & 89.9  & 20.2  & 13.6  & 90.6  & 23.7\tabularnewline
\small{+$L_{\text{Ent}}$ (space A)} & 35.7  & 66.0  & 46.3  & 39.4  & 75.5  & 51.7\tabularnewline
\small{+$L_{\text{Ent}}$ (space B)}  & 37.8  & 67.0  & 48.3  & 41.1  & 80.2  & 54.2\tabularnewline
\small{+$L_{\text{Ent}}$ (space C)} & 39.8 & 70.1 & 50.8 & 46.2 & 71.6 & 56.2\tabularnewline
\small{+$L_{\text{MI}}$ (space C)} & 39.3  & 72.9  & 51.1  & 49.5  & 70.9  & 58.2\tabularnewline
\small{+$L_{\text{MI}}$+$L_{\text{EC}}$ } & 45.2  & 72.6  & 55.7  & 52.7  & 74.1  & \textbf{61.6}\tabularnewline
\small{+$L_{\text{MI}}$+$L_{\text{EC}}$+$L_{\text{SPCE}}$} & 50.2  & 71.5  & \textbf{59.0}  & 52.2 & 74.3 & 61.3\tabularnewline
\hline 
\end{tabular}\label{all-tabel1}
\end{table}

\subsection{Conventional zero shot learning results}
We investigate the proposed method for conventional ZSL that only recognize unseen classes at the test stage. And we compare the result of our
method with several state of the art results from recent works \cite{Akata_etal_2015EOEFGIC}\cite{Paredes_Toor_2015AESATZs}\cite{Changpinyo_etal_2016SCZs}\cite{Annadani_Biswas_2018PSRZs}\cite{Tong2019HierarchicalDO}.
As shown in \cref{c-table}, the proposed approach compares favorably with the existing
approaches in literature, obtaining the state-of-the-art on SUN, AwA2
and aPY datasets. On CUB dataset, our result is 2\% lower than DLFZRL \cite{Tong2019HierarchicalDO}.

\begin{table}
\centering
\caption{Results of Conventional Zero-Shot Learning.}
\begin{tabular}{l c c c c}
\hline 
Method  & SUN  & CUB  & AwA2  & aPY\tabularnewline
\hline 
DAP\cite{Lampert_etal_2014AbCZsVBC}  & 39.9  & 40.0  & 46.1  & 33.8\tabularnewline
CONSE\cite{Mohammad_etal_2014ZsLCCSE}  & 38.8  & 34.3  & 44.5  & 26.9\tabularnewline
ALE\cite{Akata_etal_2013LeAbC}  & 58.1  & 54.9  & 62.5  & 39.7\tabularnewline
DEVISE\cite{Frome_etal_2013DEVISE}  & 56.5  & 52.0  & 59.7  & 39.8\tabularnewline
SJE\cite{Akata_etal_2015EOEFGIC}  & 53.7  & 53.9  & 61.9  & 32.9\tabularnewline
ESZSL\cite{Paredes_Toor_2015AESATZs}  & 54.5  & 53.9  & 58.6  & 38.3\tabularnewline
SYNC\cite{Changpinyo_etal_2016SCZs}  & 40.3  & 55.6  & 46.6  & 23.9\tabularnewline
PSR\cite{Annadani_Biswas_2018PSRZs}  & 61.4  & 56.0  & 63.8  & 38.4\tabularnewline
DLFZRL\cite{Tong2019HierarchicalDO}  & 59.3  & \textbf{57.8}  & 63.7  & 44.5\tabularnewline
\textbf{Proposed}  & \textbf{62.1}  & 57.6  & \textbf{64.6}  & \textbf{44.7}\tabularnewline
\hline
\end{tabular}\label{c-table}
\end{table}

\begin{table*}[htbp]
\centering
\caption{Results of Generalized Zero-Shot Learning on four datasets under Proposed
Splits (PS)\cite{Xian_etal_2017ZsTGTBTU}}
\small{
\begin{tabular}{l|ccc|ccc|ccc|ccc} \hline
\multirow{2}{1cm}{\centering Methods} & \multicolumn{3}{c|}{AwA2} & \multicolumn{3}{c|}{CUB} 
& \multicolumn{3}{c|}{SUN} & \multicolumn{3}{c}{aPY} \\
\cline{2-13}
~ & {\centering ts} & {\centering tr} & {\centering H}
& {\centering ts} & {\centering tr} & {\centering H}
& {\centering ts} & {\centering tr} & {\centering H}
& {\centering ts} & {\centering tr} & {\centering H} \\
\hline
\textbf{Non-Generative Models}  &  &  &  &  &  &  &  &  &  &  &  & \\
ALE\cite{Akata_etal_2013LeAbC}  & {16.8 }  & {76.1 }  & {27.5 }  & {23.7 }  & {62.8 }  & {34.4 }  & {21.8 }  & {33.1 }  & {26.3 }  & {4.6 }  & {73.7 }  & {8.7}\tabularnewline
DeViSE\cite{Frome_etal_2013DEVISE}  & {13.4 }  & {68.7 }  & {22.4 }  & {23.8 }  & {53.0 }  & {32.8 }  & {16.9 }  & {27.4 }  & {20.9 }  & {4.9 }  & {76.9 }  & {9.2}\tabularnewline
SynC\cite{Changpinyo_etal_2016SCZs}  & {8.9 }  & {87.3 }  & {16.2 }  & {11.5 }  & {70.9 }  & {19.8 }  & {7.9 }  & {43.3 }  & {13.4 }  & {7.4 }  & {66.3 }  & {13.3}\tabularnewline
ZSKL\cite{Zhang_etal_2018ZsKL}  & {18.9 }  & {82.7 }  & {30.8 }  & {21.6 }  & {52.8 }  & {30.6 }  & {20.1 }  & {31.4 }  & {24.5 }  & {10.5 }  & {76.2 }  & {18.5}\tabularnewline
DCN \cite{Liu_etal_2018GZsDCN}  & {25.5 }  & {84.2 }  & {39.1 }  & {28.4 }  & {60.7 }  & {38.7 }  & {25.5 }  & {37.0 }  & {30.2 }  & {14.2 }  & {75.0 }  & {23.9}\tabularnewline
DLFZRL \cite{Tong2019HierarchicalDO}  & {- }  & {- }  & {45.1 }  & {- }  & {- }  & {37.1 }  & {- }  & {- }  & {24.6 }  & {- }  & {- }  & {31.0}\tabularnewline
\textbf{Proposed}  & 52.7  & 74.1  & \textbf{61.6} & {40.6 }  & {55.1 }  & \textbf{46.7}{ }  & 41.7  & 37.4  & \textbf{39.5}  & 31.5  & 51.8  & \textbf{39.2}\tabularnewline
\hline 
\textbf{Generative Models}  &  &  &  &  &  &  &  &  &  &  &  & \tabularnewline
f-CLSWGAN\cite{Xian_etal_2018FGNZs}  & {52.1 }  & {68.9 }  & {59.4 }  & {43.7 }  & {57.7 }  & {49.7 }  & {42.6 }  & {36.6 }  & {39.4 }  & {- }  & {- }  & {-}\tabularnewline
F-VAEGAN-D2\cite{Xian2019FVAEGAND2AF}  & {57.6 }  & {70.6 }  & {63.5 }  & {48.4 }  & {60.1 }  & 53.6  & {45.1 }  & {38.0 }  & {41.3 }  & {- }  & {- }  & {-}\tabularnewline
CADA-VAE\cite{Schonfeld_etal_2019GZsAVA}  & {55.8 }  & {75.0 }  & {63.9 }  & {51.6 }  & {53.5 }  & {52.4 }  & {47.2 }  & {35.7 }  & {40.6 }  & {- }  & {- }  & {-}\tabularnewline
CRnet\cite{Zhang2019CoRepresentationNF}  & {52.6 }  & {52.6 }  & {63.1 }  & {45.5 }  & {56.8 }  & {50.5 }  & {34.1 }  & {36.5 }  & {35.3 }  & {32.4 }  & {68.4 }  & {44.0}\tabularnewline
DLFZRL+softmax\cite{Tong2019HierarchicalDO}  & {- }  & {- }  & {60.9 }  & {- }  & {- }  & {51.9 }  & {- }  & {- }  & 42.5  & {- }  & {- }  & {38.5}\tabularnewline
TCN \cite{Jiang2019TransferableCN}  & 61.2  & 65.8  & 63.4  & 52.6  & 52.0  & 52.3  & 31.2  & 37.3  & 34.0  & 24.1  & 64.0  & 35.1\tabularnewline
GDAN \cite{Huang2019GenerativeDA}  & 32.1  & 67.5  & 43.5  & 39.3  & 66.7  & 49.5  & 38.1  & 89.9  & 53.4  & 30.4  & 75.0  & 43.4\tabularnewline
IZF \cite{Shen2020InvertibleZR}  & 60.6  & 77.5  & 68.0  & 52.7  & 68.0  & 59.4  & 52.7  & 57.0  & \textbf{54.8}  & 42.3  & 60.5  & \textbf{49.8}\tabularnewline
DVBE \cite{Min2020DomainAwareVB}  & 62.7  & 77.5  & 69.4  & 64.4  & 73.2  & \textbf{68.5}  & 44.1  & 41.6  & 42.8  & 37.8  & 55.9  & 45.2\tabularnewline
IAS \cite{Chou2021AdaptiveAG}  & 65.1  & 78.9  & \textbf{71.3}  & 41.4  & 49.7  & 45.2  & 29.9  & 40.2  & 34.3  & 35.1  & 65.5  & 45.7\tabularnewline
\textbf{f-CLSWGAN+Proposed}  & 56.4  & 83.2  & 67.2  & 52.1  & 55.8  & 53.9  & 53.3  & 35.0  & 42.3  & 37.1  & 57.7  & 45.2\tabularnewline
\hline
\end{tabular}\label{all-table}}
\end{table*}

\subsection{Generalized zero shot learning results}
\subsubsection*{Comparison with deterministic models}
We compare the performance of our proposed model with several recent deterministic models for GZSL. Taking DCN \cite{Liu_etal_2018GZsDCN} as the baseline model, as shown in \cref{all-table}, our method gains a superior accuracy compared to other deterministic models on all datasets: it
obtains $21\%\thicksim64\%$ improvements over DCN and significantly outperforms
a previous state-of-art deterministic model-DLFZRL \cite{Tong2019HierarchicalDO}.
Besides, we observe that our deterministic model obtains
superior results than some generative models, for example, it outperforms the f-CLSWGAN on all datasets.

\subsubsection*{Comparison with generative models}
We investigate the performance of our proposed method by incorporating a generative model, f-CLSWGAN. Seen image features and class-level attributes are used to train f-CLSWGAN
and image features of unseen classes can be generated by unseen
class-level attributes. Including the generated features for target
classes into the entire training, we train the model by the loss function defined in \cref{sec:generative}.
As shown in \cref{all-table}, our proposed model significantly
outperforms the baseline model f-CLSWGAN with $7\%\thicksim 13\%$ improvements. Our proposed method achieve comparable results with several recently
proposed sophisticated generative models. Moreover, unlike some generative
models, such as \cite{Jiang2019TransferableCN,Huang2019GenerativeDA,Chou2021AdaptiveAG} gain superior results on one or two datasets but inferior results
on other dataset,  our proposed method obtain favorable results on all the dataset: it  ranks as the top 3 $\thicksim$  top 5 method for each dataset.

\section{Conclusion}
This paper addresses information-theoretic loss functions to quantify the knowledge transfer and semantic relation for GZSL/ZSL. Leveraging
on the proposed probability vector representation based on the prototypical model, the proposed loss can be effectively evaluated with simple closed forms. Experiments show that our approach yields state of the art performance for deterministic approach for GZSL and conventional ZSL tasks. Moreover, by incorporating with generated data from f-CLSWGAN, the proposed method also gains favorable performance. One limitation of this work is that we need pretty much extra cross validation to select the hyperparameters; Another limitation is that the loss functions have certain correlation, so further study is needed to simply the loss functions while keeping similar performance, and reduce the number of hyperparameters.

\clearpage
{\small{}{}{}{}{}  \bibliographystyle{ieee_fullname}
\bibliography{zsl_ref}
 }{\small\par}

\end{document}